\documentclass[conference]{IEEEtran}
\IEEEoverridecommandlockouts
\usepackage{booktabs}
\usepackage{cite}
\usepackage{amsmath,amssymb,amsfonts}
\usepackage{algorithmic}
\usepackage{graphicx}
\usepackage{textcomp}
\usepackage{xcolor}
\usepackage[caption=false,font=footnotesize]{subfig}
\usepackage{soul}
\usepackage{hyperref} 
\usepackage{balance}

\def\BibTeX{{\rm B\kern-.05em{\sc i\kern-.025em b}\kern-.08em
    T\kern-.1667em\lower.7ex\hbox{E}\kern-.125emX}}
\begin{document}

\title{LLM-Generated Fault Scenarios for Evaluating Perception-Driven Lane Following in Autonomous Edge Systems\\
{}
}

\author{\IEEEauthorblockN{1\textsuperscript{st} Faezeh Pasandideh}
\IEEEauthorblockA{\textit{Department of Electrical Engineering} \\
\textit{Hamm-Lippstadt University. of Applied
Sciences (HSHL)}\\
Lippstadt, Germany \\
faezeh.pasandideh@hshl.de}
\and
\IEEEauthorblockN{2\textsuperscript{nd} Achim Rettberg}
\IEEEauthorblockA{\textit{Department of Electrical Engineering} \\
\textit{Hamm-Lippstadt University. of Applied
Sciences (HSHL)}\\
Lippstadt, Germany \\
achim.rettberg@hshl.de }
}

\maketitle
\begin{abstract}
 Deploying autonomous vision systems on edge devices faces a critical challenge: resource constraints prevent real-time and predictable execution of comprehensive safety tests. Existing validation methods depend on static datasets or manual fault injection, failing to capture the diverse environmental hazards encountered in real-world deployment. To address this, we introduce a decoupled offline–online fault injection framework. This architecture separates the validation process into two distinct phases: a computationally intensive Offline Phase and a lightweight Online Phase. In the offline phase, we employ Large Language Models (LLMs) to semantically generate structured fault scenarios and Latent Diffusion Models (LDMs) to synthesize high-fidelity sensor degradations. These complex fault dynamics are distilled into a pre-computed lookup table, enabling the edge device to perform real-time fault-aware inference without running heavy AI models locally. We extensively validated this framework on a ResNet18 lane-following model across 460 fault scenarios.Results show that while the model achieves a baseline $R^2 \approx 0.85$ on 
clean data, our generated faults expose significant robustness degradation---with 
RMSE increasing by up to 99\% and within-0.10 localization accuracy dropping to 
as low as 31.0\% under fog conditions---demonstrating the inadequacy of 
normal-data evaluation for real-world edge AI deployment.
\end{abstract}

\begin{IEEEkeywords}
Fault Injection, Autonomous Vehicles, Edge Computing, Large Language Models (LLMs), Perception Systems, Lane Following, Robustness Evaluation
\end{IEEEkeywords}

\section{Introduction}

Autonomous vehicles and edge-deployed robotic systems increasingly rely on 
AI-based vision pipelines for safety-critical tasks such as lane following and 
obstacle detection. While these systems have demonstrated remarkable 
capabilities under nominal conditions, their robustness under degraded visual 
inputs — caused by sensor faults, adverse weather, lighting changes, or 
hardware failures — remains a fundamental open challenge. Ensuring that such 
systems behave safely under fault conditions is essential before they can be 
deployed in real-world environments.

Current approaches to safety validation suffer from several limitations. 
Reactive fault detection methods identify failures only after they occur, 
providing no mechanism for proactive evaluation or early warning. Manual 
fault injection is labor-intensive, difficult to scale, and offers limited coverage 
of the vast space of possible fault conditions. Real-world testing under 
adversarial or degraded conditions is both costly and dangerous, particularly 
when edge-deployed hardware such as NVIDIA Jetson is involved. Furthermore, 
state-of-the-art generative AI models — including Large Language Models 
(LLMs) and Latent Diffusion Models (LDMs) — demand computational resources far exceeding the memory and processing 
capacity of edge devices, making their direct deployment on such platforms 
impractical for real-time systems.

Existing research has explored these challenges in isolation. LLM-based 
frameworks have been used to generate behavioral fault scenarios 
\cite{11047235, 11229755}, while perception-focused benchmarks have 
evaluated lane detection robustness under predefined visual perturbations 
\cite{10.1145/3664647.3680761, 10.1145/3180155.3180220}. Control-theoretic 
approaches have analyzed stability under parametric uncertainty \cite{7890875}. 
However, none of these works unify semantic fault generation, realistic image 
synthesis, and edge hardware evaluation under real-time constraints into a single framework, leaving a 
critical gap between advanced AI-driven testing methodologies and practical 
autonomous driving deployment.

To address these limitations, we propose a decoupled virtual testing framework 
with a two-phase architecture. The core 
insight is that resource-intensive AI models need not run on the edge device 
itself — they can be executed offline in a cloud or high-performance computing 
environment to generate fault scenarios and precompute predictions, while 
the edge device performs lightweight lookup table queries at runtime to assess 
fault conditions with bounded latency in real time. This separation enables comprehensive, 
AI-driven safety evaluation without compromising the deployment constraints 
of resource-limited edge hardware.

In the offline phase, LLMs are used to generate semantically rich 
fault scenario descriptions, LDMs synthesize corresponding faulty images 
that simulate realistic visual degradations, and VLMs validate the generated 
scenarios and predict lane-following performance under each fault condition. 
The outputs of this phase are stored in a structured lookup table. In the 
online phase, the autonomous system is deployed on an NVIDIA Jetson 
edge device, which queries the precomputed lookup table to evaluate fault 
conditions in real time with minimal runtime overhead without executing any computationally expensive AI 
models locally. This hybrid architecture effectively bridges the gap between 
advanced generative AI capabilities and practical edge deployment feasibility.

The main contributions of this paper are as follows:
\begin{itemize}
    \item A decoupled offline–online framework integrating LLM-based 
    fault specification and LDM-based faulty image synthesis for 
    proactive safety validation of perception-driven lane-following 
    systems under real-time constraints.
    
    \item A lookup-table-based online inference mechanism enabling 
    real-time fault condition assessment on resource-constrained edge 
    devices without executing generative AI models at runtime.
    
    \item Comprehensive evaluation of lane-following robustness under 
    semantically diverse fault scenarios on NVIDIA Jetson-based edge 
    platforms, including analysis of real-time performance and resource utilization.
\end{itemize}

The remainder of this paper is organized as follows. Section~\ref{sec:related} 
reviews related work on fault injection, adversarial scenario generation, and 
lane detection robustness. Section~\ref{sec:method} describes the proposed 
decoupled framework in detail, covering both the offline and online phases. 
Section~\ref{sec:experiments} presents the experimental setup, dataset, and 
evaluation metrics. 
Section~\ref{sec:conclusion} concludes the paper and outlines directions 
for future work.

\section{Related Work}
\label{sec:related}
Authors in \cite{11047235} propose LLM-Attacker, a closed-loop adversarial 
scenario generation framework that leverages multiple collaborative LLM agents 
to analyze complex traffic scenes and identify adversarial vehicles whose 
trajectories are optimized to create dangerous interactions with the ego vehicle. 
Evaluated on the Waymo Open Motion Dataset within the MetaDrive simulation 
environment, the framework employs LLaMA 3.1 (8B parameters) through 
iterative initialization, reflection, and modification modules, achieving higher 
attack success rates than baselines such as random selection and minimum 
time-to-collision methods. Training an autonomous driving system on these 
generated scenarios reduced collision rates by approximately 50\% compared 
to training with normal scenarios. However, the study focuses on behavior-level 
adversarial interactions rather than perception-level faults, leaving a gap for 
research that integrates LLM-generated fault descriptions with diffusion-based 
image synthesis to evaluate lane-following robustness under degraded visual 
conditions, particularly under real-time execution constraints.

Authors in \cite{11229755} propose LOFT, a two-stage LLM pipeline in which 
the first LLM converts structured simulation data into natural language 
descriptions and recommends potential fault types, while the second analyzes 
scenario context to identify high-risk time intervals. These outputs initialize a 
multi-objective genetic search algorithm that explores fault parameters including 
type, timing, duration, and deviation magnitude. Evaluated in simulation using 
an Apollo-like system across six driving scenarios, LOFT injects 17 fault types 
across five modules — localization, perception, prediction, planning, and 
control — using GPT-4o, detecting over 90\% more critical faults than random 
and DBN-based baselines. However, the framework targets system-level faults 
rather than perception-level visual degradations and does not employ generative 
models such as GANs or latent diffusion models for synthetic image synthesis, 
leaving a gap for frameworks that integrate LLM-generated fault descriptions 
with diffusion-based image synthesis for lane-following evaluation, with 
deployment on resource-constrained edge platforms.

Authors in \cite{10.1145/3664647.3680761} introduce LanEvil, a benchmark for 
evaluating the robustness of deep learning-based lane detection systems under 
naturally occurring visual perturbations such as shadows, reflections, road 
cracks, tire marks, and traffic obstructions. Using the CARLA simulator, 94 
customizable 3D scenarios were created to synthesize 90,292 images covering 
14 illusion types across multiple severity levels, with evaluation conducted on 
models including LaneATT, SCNN, UltraFast, GANet, and BezierLaneNet using 
accuracy and F1-score metrics. Results reveal an average drop of 5.37\% in 
accuracy and 10.70\% in F1-score, with shadow-based illusions causing the 
largest degradation, and tests on real systems such as OpenPilot and Apollo 
confirm that such illusions can lead to incorrect perception decisions and 
potential collisions. However, the benchmark relies on predefined visual 
perturbations without incorporating LLMs for fault scenario generation or 
generative models such as latent diffusion models for image synthesis, and 
does not consider real-time constraints during deployment.

Authors in \cite{10.1145/3180155.3180220} propose DeepTest, an automated 
testing framework that evaluates the robustness of DNN-based autonomous 
driving systems by systematically applying image transformations — including 
brightness changes, blurring, fog, and rain effects — to simulate real-world 
visual disturbances, using neuron coverage as a testing metric and metamorphic 
relations to detect erroneous behavior. Evaluated on end-to-end driving models 
such as Rambo, Chauffeur, and Epoch using the Udacity dataset, the framework 
detected thousands of erroneous behaviors and improved MSE by up to 46\% 
after retraining under adverse conditions. However, the fault generation relies 
on predefined transformations that lack semantic richness, and no LLM or 
generative model such as a latent diffusion model is involved, nor is real-time 
execution on edge hardware addressed.

Authors in \cite{7890875} develop a mathematical vehicle dynamics model 
combined with a Model Predictive Controller to analyze how parametric 
uncertainties — specifically road–tire friction coefficient and camera look-ahead 
distance — affect lane-following stability in closed-loop MATLAB/Simulink 
simulation. The controller maintains lateral and angular errors within acceptable 
bounds despite road curvature disturbances; however, the study relies on a 
control-based model rather than deep learning, and does not consider visual 
perception faults such as image blur, occlusion, or brightness changes, nor does 
it explore LLM-generated fault scenarios or diffusion-based image synthesis, 
or their implications under real-time constraints.

Despite significant progress in autonomous driving robustness, existing approaches remain fragmented. LLM-based methods primarily focus on behavior- or system-level faults, while perception-focused studies rely on predefined or manually designed visual perturbations lacking semantic richness. Furthermore, most evaluations are confined to simulation and do not consider deployment on resource-constrained edge devices with real-time performance constraints. Additionally, generative models such as latent diffusion models have not been leveraged to synthesize realistic fault scenarios for lane-following evaluation. Consequently, a unified framework that integrates LLM-based fault specification, diffusion-based image synthesis, and real-time lane-following evaluation on edge platforms is still missing, representing a critical gap that this paper addresses.

\section{Methodology}
\label{sec:method}

\begin{figure}[htbp]
\centerline{\includegraphics[width=\columnwidth]{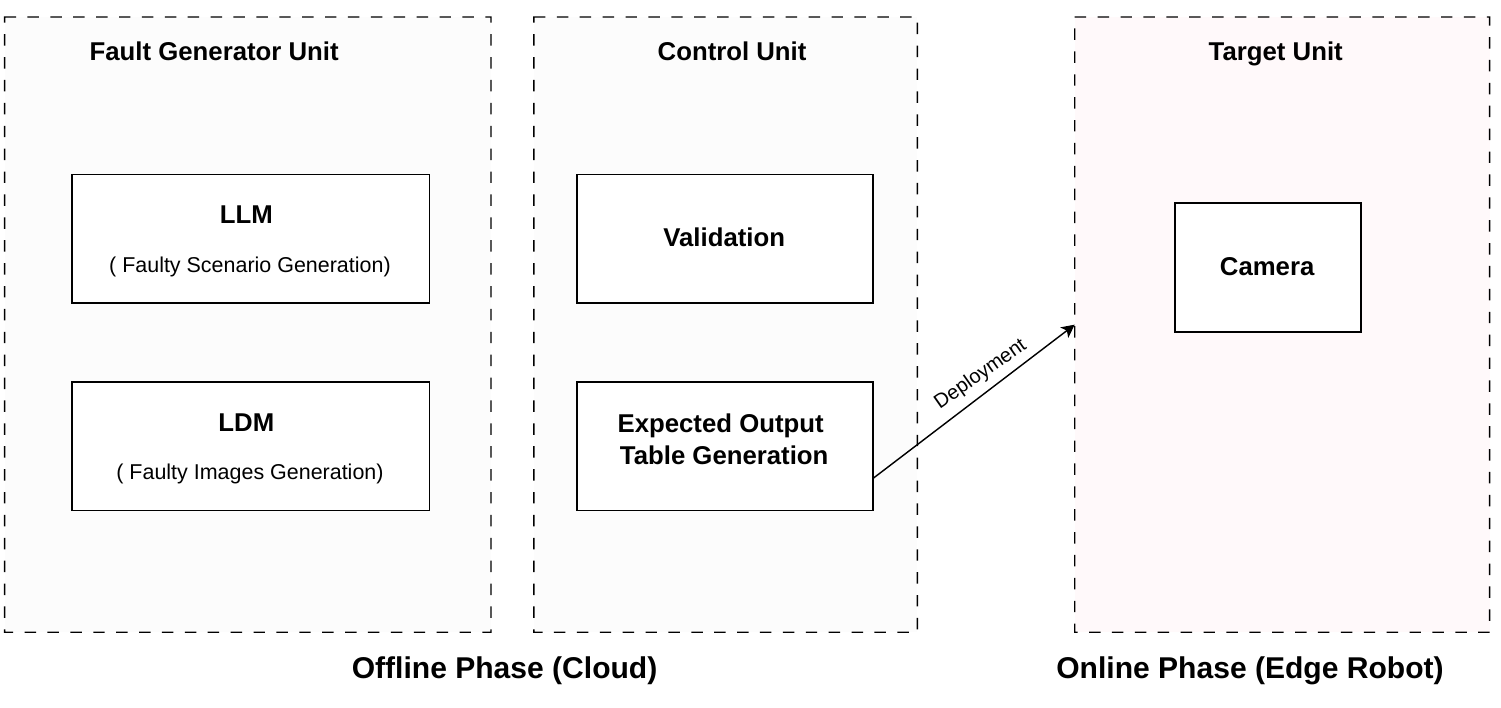}}
\caption{ Proposed decoupled framework architecture}
\label{fig:off-on}
\end{figure}

The proposed framework, illustrated in Figure~\ref{fig:off-on}, is built around 
a practical observation: the most computationally demanding steps of fault 
generation and validation do not need to happen on the robot itself. By 
separating the heavy offline computation from the lightweight online deployment, 
the framework makes it possible to use state-of-the-art generative AI models 
for thorough safety testing while still running the final system on a 
resource-constrained edge device such as the NVIDIA Jetson Nano.
The offline pipeline consists of three tightly coupled components. The first is 
the Semantic Scenario Generator, which uses an LLM to produce 
structured, natural-language descriptions of fault conditions that a camera 
sensor might encounter in real-world autonomous driving, for example, 
lens blur caused by rain accumulation, overexposure under direct sunlight, 
or partial occlusion from road debris. Rather than relying on a fixed set of 
hand-crafted perturbations, the LLM draws on its broad knowledge to generate 
diverse and semantically coherent fault descriptions that reflect conditions 
a deployed system could plausibly face.

These textual descriptions are then passed to the Sensor Degradation 
Synthesizer, which uses a Latent Diffusion Model (LDM) to translate each 
fault description into a corresponding faulty image. The LDM operates in a 
compressed latent space, conditioning the image generation process on the 
fault description to produce realistic visual degradations of the original driving 
scene. This step is what gives the framework its ability to go beyond simple 
image filters: the synthesized images capture the complex, non-linear 
appearance of real sensor faults rather than approximating them with 
brightness adjustments or Gaussian blur.

The third component evaluates the quality and semantic fidelity of the 
generated images using CLIP-based similarity scoring. For each 
generated image, the CLIP model computes the alignment between the visual 
output and the original fault description, providing a quantitative measure of 
whether the LDM has faithfully rendered the intended degradation. Images 
that fall below a similarity threshold are filtered out, ensuring that only 
high-quality, semantically consistent fault samples are retained in the dataset.

The outputs of these three components, the fault descriptions, the synthesized 
images, and their associated CLIP scores, are stored in a structured lookup 
table. During online deployment, the edge device does not re-run any of these 
models. Instead, it queries the precomputed table to retrieve the relevant fault 
assessment for the current operating condition, enabling real-time safety 
evaluation within the tight latency and memory budget of the Jetson Nano.

\subsection{Dataset}
\begin{figure}[htbp]
    \centering
    \includegraphics[width=0.65\linewidth]{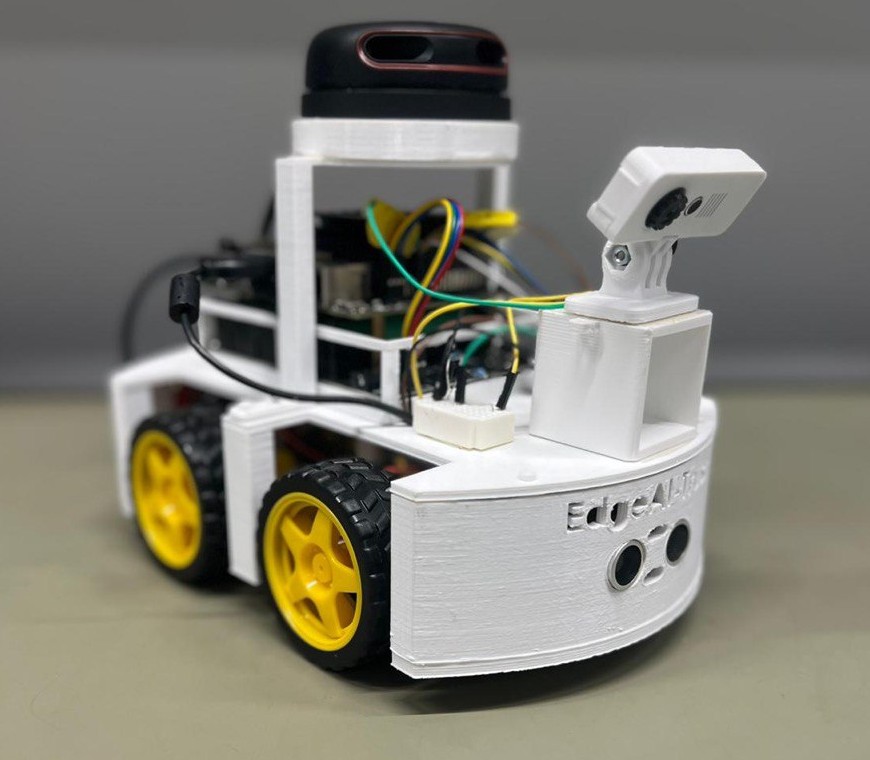}
    \caption{The robot platform used for edge inference experiments.}
    \label{fig:JetBot}
\end{figure}

Real-world data were collected using the NVIDIA JetBot platform, shown in Figure \ref{fig:JetBot} on a 
physical track, yielding 796 RGB images ($224 \times 224$ pixels) across 
lane-following and obstacle detection tasks. Since this volume is insufficient 
for evaluating models under diverse fault conditions, we augmented these 
recordings into VisionFault-350K~\cite{azarafza_2026_18695332}, a 
fault-injected dataset of 350,751 images, using the offline phase of our 
framework described below.

\subsection{Scenario Generator (LLM)}
GPT-OSS~\cite{gpt-oss} was used to generate approximately 10,000 fault 
scenario descriptions covering categories such as camera failures, motion 
blur, extreme weather (fog, rain, ice), low-light conditions, and lens 
distortions. The full scenario list is available in our \href{https://anonymous.4open.science/r/LLM-Based-Generation-and-Evaluation-of-Fault-Scenarios-for-Autonomous-Vehicle-Edge-Computing-1FFF/SyntheticScenarios/gpt_oss_scenarios.txt}{GitHub repository}.
\subsection{Sensor Fault Synthesis (LDM)}
\begin{figure}[htbp]
\centerline{\includegraphics[width=0.7\columnwidth]{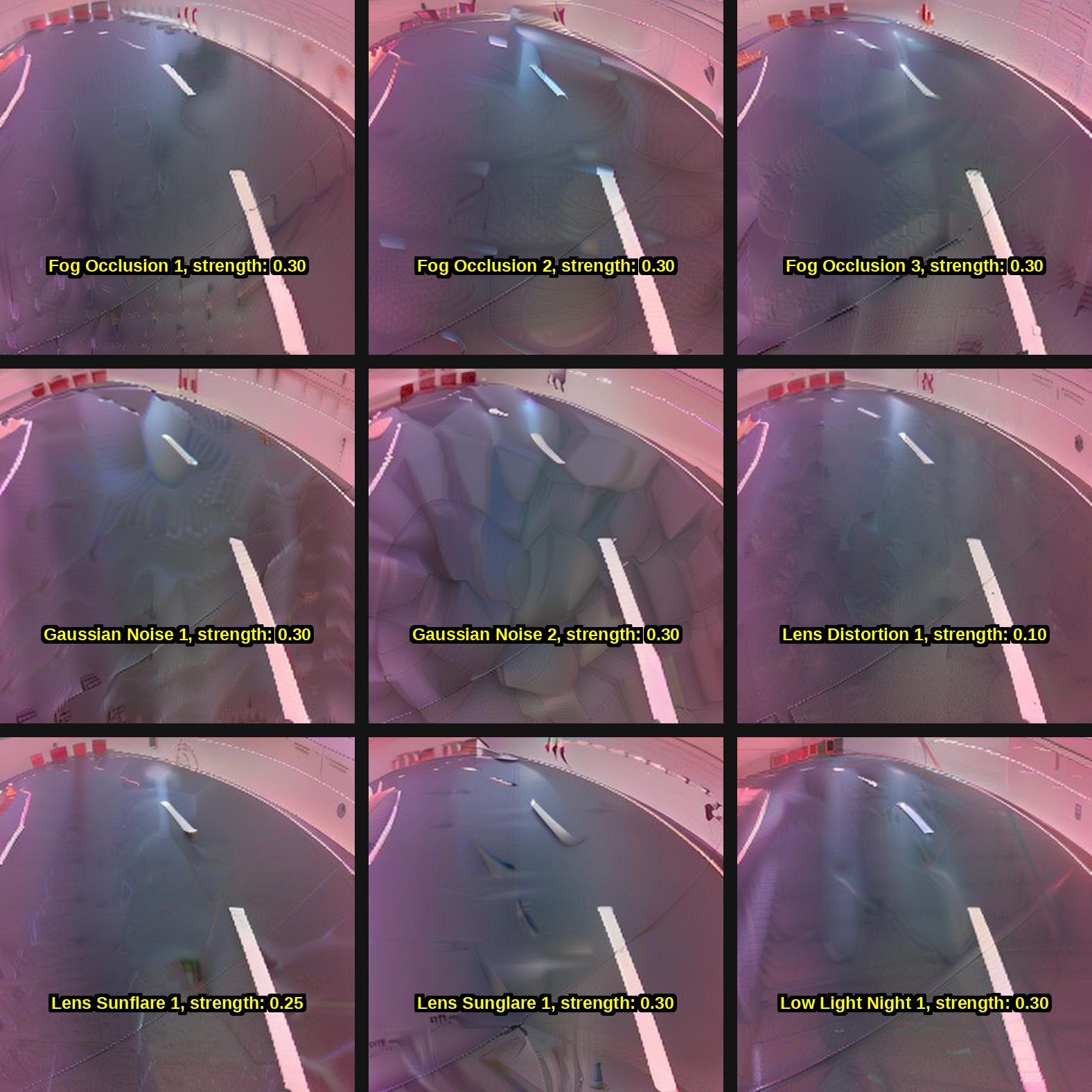}}
\caption{LDM-generated images using LLM-derived denoising strengths.}
\label{fig:LDM}
\end{figure}

Each LLM-generated description was used to condition Stable Diffusion 
2.1~\cite{stable-fi} in image-to-image mode, synthesizing a degraded 
variant $I'_{fault}$ of the original frame while preserving its underlying 
scene structure. The degree of visual corruption is controlled by the 
denoising strength parameter, whose value is derived directly from the 
LLM output for each scenario. Figure~\ref{fig:LDM} shows nine 
representative examples generated across varying strength values.

\section{Evaluation}
\label{sec:experiments}
Our evaluation validates the framework's capacity to generate realistic, high-impact faults and expose fragile performance in the target model. We used a standard ResNet18 regression backbone for lane-following as the perception stack.
\subsection{Semantic Fidelity Validation(VLM/CLIP)}

Prior to performance testing, we ensured the integrity of the generated fault images. We employed CLIP (ViT-L/14)\cite{clip}, to measure the semantic consistency between the LLM's textual prompt $T(S_i)$ and the LDM's synthesized image $I'_{fault}$. This is a critical filtering step, as it guarantees that the injected faults are not random noise but are semantically aligned with the intended hazard. Low-fidelity generations failing a predetermined similarity threshold.

\subsection{Lane-following Performance: ResNet18 on normal data}

\begin{figure*}[t]
    \centering
    \includegraphics[width=\linewidth]{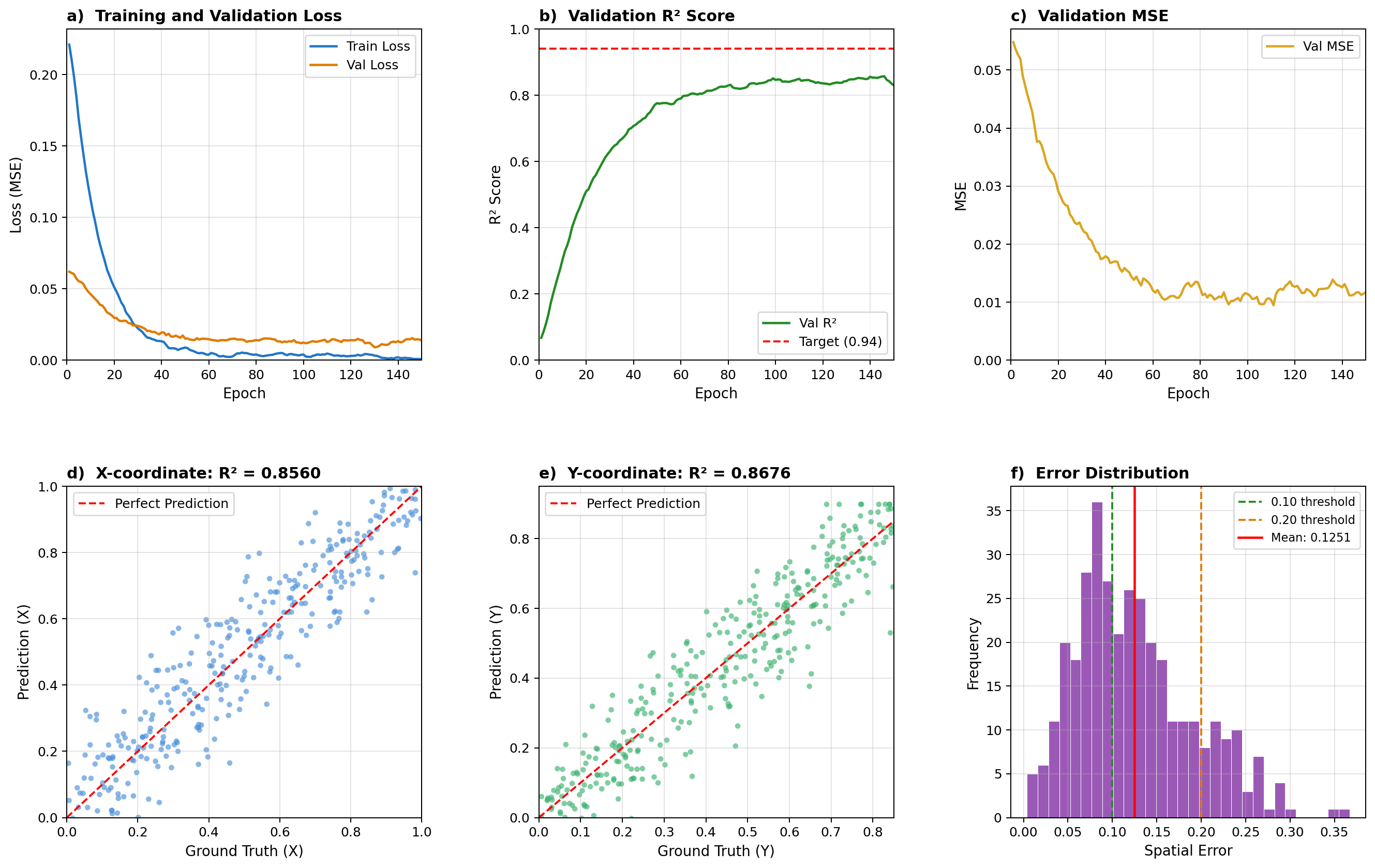}
    \caption{ResNet-18 performance on normal lane-following data (150 epochs):
    (a) train/val loss converging to ${\sim}0.011$, (b) validation $R^2$
    reaching ${\sim}0.85$ (target: 0.94), (c) MSE declining to ${\sim}0.011$,
    (d) X-coordinate scatter ($R^2\!=\!0.8560$), (e) Y-coordinate scatter
    ($R^2\!=\!0.8676$), and (f) spatial error distribution
    (mean\,=\,0.1251).}
    \label{fig:resnet18_normal}
\end{figure*}

Fig.~\ref{fig:resnet18_normal} presents the training dynamics and prediction
performance of the ResNet-18 model trained on normal (fault-free)
lane-following data over 150 epochs, employing SmoothL1 loss, partial layer
freezing, and a learning rate scheduler. As shown in
Fig.~\ref{fig:resnet18_normal}(a), the training and validation loss curves
demonstrate rapid convergence within the first 20 epochs, with the training
loss declining sharply from approximately 0.22 to below 0.001 and stabilizing
thereafter; the validation loss plateaus near 0.011, indicating a mild but
stable generalization gap without severe overfitting. The validation $R^2$
score in Fig.~\ref{fig:resnet18_normal}(b) exhibits a steep upward trend
within the first 30 epochs, followed by continued smooth improvement,
ultimately converging to approximately 0.85 by epoch 150; the score remains
below the target threshold of 0.94 (dashed red line), reflecting the inherent
difficulty of precise lane-center regression under limited training data.
Correspondingly, the validation MSE in Fig.~\ref{fig:resnet18_normal}(c)
declines sharply from approximately 0.055 in the initial epochs, stabilizing
near 0.011 by epoch 150 with minor oscillations after epoch 80, reflecting
steady and sustained improvement in prediction accuracy.

At the per-coordinate level, the scatter plots in
Figs.~\ref{fig:resnet18_normal}(d) and~\ref{fig:resnet18_normal}(e) illustrate
prediction quality for the X- and Y-coordinates independently, with the model
achieving $R^2 = 0.8560$ and $R^2 = 0.8676$, respectively; predictions cluster
closely around the perfect-prediction diagonal across the full normalized range,
with the Y-coordinate exhibiting a slight concentration of predictions near
ground-truth values around 0.5, while both coordinates display a moderate spread
of outlier predictions at extreme values, consistent with challenging or ambiguous
frames in the validation set. Finally, the spatial error distribution in
Fig.~\ref{fig:resnet18_normal}(f) reveals that the majority of predictions fall
within a spatial error below 0.10 (green dashed threshold), with a mean spatial
error of 0.1251 (red solid line); the distribution peaks between 0.05 and 0.12
with a moderate tail extending to approximately 0.35, establishing the baseline
performance of the model prior to fault injection and motivating the need for
robustness evaluation under LLM-LDM-generated degradations.

\subsection{Lane-Following Performance of ResNet-18 on LLM-LDM-Generated Fault-Injected Data}
\begin{figure}[!t]
    \centering
    \includegraphics[width=\linewidth]{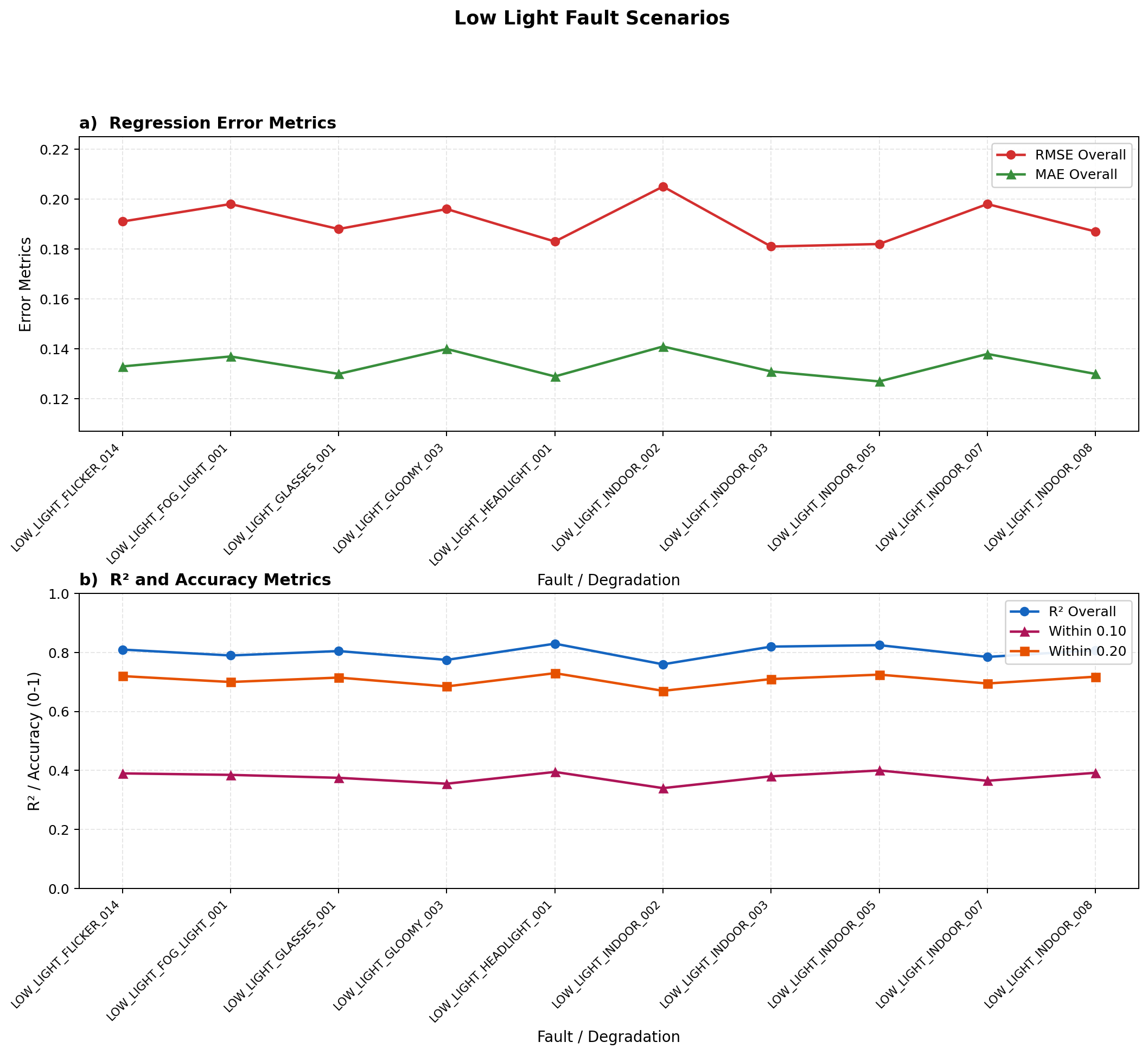}
    \caption{ResNet18 performance on LDM-generated faulty images:
    low-light fault scenarios. Top: regression error metrics (RMSE and MAE).
    Bottom: goodness-of-fit and accuracy metrics ($R^2$, Within~0.10, Within~0.20).}
    \label{fig:fault_lowlight}
\end{figure}

\begin{figure}[!t]
    \centering
    \includegraphics[width=\linewidth]{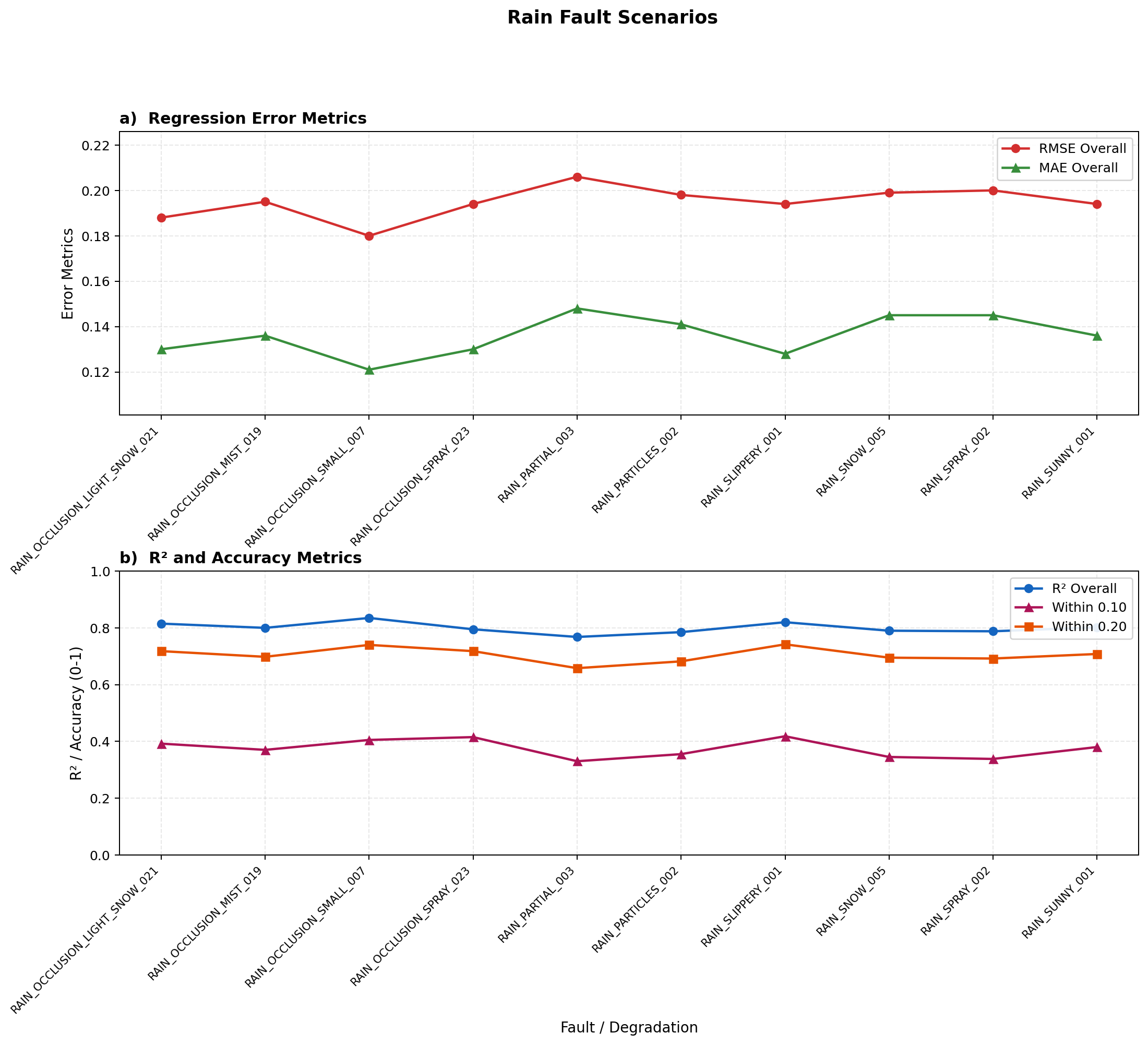}
    \caption{ResNet18 performance on LDM-generated faulty images:
    rain-related fault scenarios. Top: regression error metrics (RMSE and MAE).
    Bottom: goodness-of-fit and accuracy metrics ($R^2$, Within~0.10, Within~0.20).}
    \label{fig:fault_rain}
\end{figure}

\begin{figure}[!t]
    \centering
    \includegraphics[width=\linewidth]{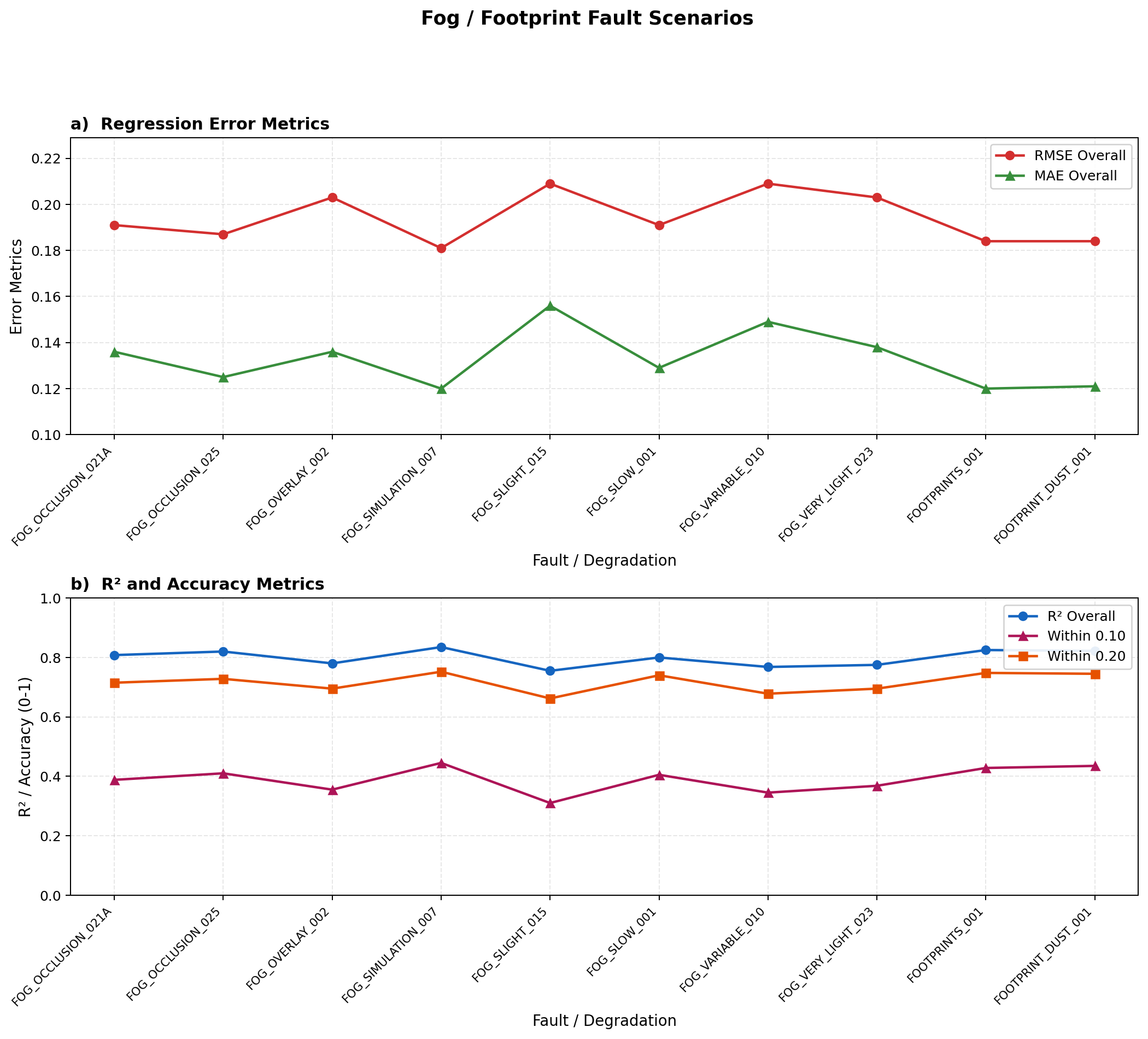}
    \caption{ResNet18 performance on LDM-generated faulty images:
    fog and footprint fault scenarios. Top: regression error metrics (RMSE and MAE).
    Bottom: goodness-of-fit and accuracy metrics ($R^2$, Within~0.10, Within~0.20).}
    \label{fig:fault_fog}
\end{figure}

To evaluate the resilience of the lane-following model under diverse
visual degradation conditions, we extracted per-folder regression and
accuracy metrics across the full VisionFault dataset, which encompasses
a rich taxonomy of fault categories including atmospheric effects
(e.g., \texttt{FOG}, \texttt{DUST\_STORM}, \texttt{FROST\_COATING},
\texttt{RAIN}), optical and lens artifacts (e.g., \texttt{LENS\_DISTORTION},
\texttt{BARREL\_DISTORTION}, \texttt{FISH\_EYE}, \texttt{LENS\_VIGNETTING},
\texttt{CHROMATIC\_ABERRATION}), sensor and hardware faults
(e.g., \texttt{DEAD\_PIXELS}, \texttt{CAMERA\_FAILURE},
\texttt{CAMERA\_BANDING}, \texttt{SENSOR\_HEAT}, \texttt{HW\_OVERHEAT}),
motion and geometric degradations (e.g., \texttt{MOTION\_BLUR},
\texttt{CAMERA\_SHAKE}, \texttt{CAMERA\_YAW}, \texttt{PERSPECTIVE\_DISTORTION}),
and illumination-based faults (e.g., \texttt{GLARE\_OCCLUSION},
\texttt{LOW\_LIGHT\_TUNNEL}, \texttt{BRIGHT\_REFLECTION},
\texttt{COLOR\_SHIFT\_NIGHT}). Given the large number of faulty folders
generated, only three representative fault categories are presented here
for clarity: low-light conditions (Fig.~\ref{fig:fault_lowlight}),
rain-related degradations (Fig.~\ref{fig:fault_rain}), and fog/footprint
artifacts (Fig.~\ref{fig:fault_fog}).

As shown in Figs.~\ref{fig:fault_lowlight}(a)--\ref{fig:fault_fog}(a),
the aggregate error metrics reflect a measurable degradation relative to
the normal baseline across all fault types.
RMSE$_\text{Overall}$ ranges from $0.180$ to $0.209$ and
MAE$_\text{Overall}$ from $0.120$ to $0.156$ across the three subsets.
The sharpest error peaks occur under \texttt{FOG\_SLIGHT\_015} and
\texttt{FOG\_VARIABLE\_010} (RMSE\,$\approx 0.209$), \texttt{RAIN\_PARTIAL\_003}
(RMSE\,$\approx 0.206$), and \texttt{LOW\_LIGHT\_INDOOR\_002}
(RMSE\,$\approx 0.204$), while the most benign conditions are
\texttt{RAIN\_OCCLUSION\_SMALL\_007} and \texttt{FOG\_SIMULATION\_007},
yielding RMSE values as low as $0.180$--$0.181$.
The $R^2$ and within-tolerance accuracy metrics in
Figs.~\ref{fig:fault_lowlight}(b)--\ref{fig:fault_fog}(b) confirm that
the model retains partial predictive capability under synthesized faults
--- with $R^2_\text{Overall}$ largely between $0.755$ and $0.840$ ---
while the lowest $R^2$ values coincide with the highest RMSE peaks,
namely \texttt{FOG\_SLIGHT\_015} ($R^2\!\approx\!0.755$) and
\texttt{RAIN\_PARTIAL\_003} ($R^2\!\approx\!0.768$).
The within-0.20 accuracy remains relatively stable across all three
categories, ranging from $0.658$ to $0.752$, indicating that the model
preserves coarse-grained directional steering even under significant
visual degradation. The within-0.10 accuracy remains persistently low
across all conditions, ranging from $0.310$ (\texttt{FOG\_SLIGHT\_015})
to $0.445$ (\texttt{FOG\_SIMULATION\_007}), and rarely exceeding $0.42$
in the low-light and rain subsets, underscoring that fine-grained
steering precision under fault injection remains a key open challenge,
and motivating fault-aware training or domain adaptation strategies for
robust edge AI deployment.

\begin{figure}[htbp]
\centerline{\includegraphics[width=0.7\columnwidth]
{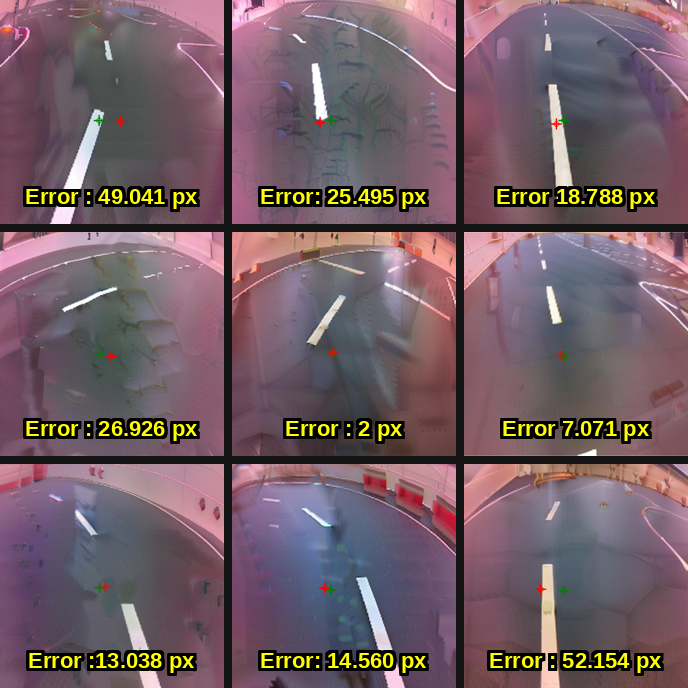}}
\caption{Prediction results for a representative fog and occlusion
degradation scenario from the VisionFault dataset, illustrating
the model's steering output under synthesized atmospheric faults.}
\label{fig:predict}
\end{figure}

Fig. \ref{fig:predict} illustrates representative prediction results of the ResNet-18 model under a fog/occlusion degradation scenario, where synthetic fog was applied to lane-following frames using the LDM-based fault injection pipeline. Each cell in the 3$\times$3 grid displays a degraded frame overlaid with the ground-truth lane-center position (green cross) and the model's predicted position (red cross), along with the corresponding pixel-level Euclidean error. The predictions exhibit a wide range of errors across the nine samples, spanning from as low as 2 px — indicating near-perfect localization despite visible fog occlusion — to as high as 52.154 px, where severe visual degradation causes the model to mislocate the lane center substantially. Intermediate error cases, such as 13.038 px, 14.560 px, and 18.788 px, reflect partial robustness where the model retains approximate spatial awareness even under moderate fog density. The high-error cases (49.041 px and 52.154 px) are visually characterized by dense fog coverage that obscures lane markings almost entirely, leaving the model with insufficient texture and edge cues for accurate regression. 
\subsection{Comparative Analysis: Normal vs.\ Fault-Injected Data}

Table~\ref{tab:comparison} summarizes the performance of ResNet-18
across normal and fault-injected conditions drawn from the VisionFault
dataset, spanning representative fault folders across atmospheric,
optical, sensor, motion, and illumination degradation families.
The model achieves a baseline $R^2$ of $\sim$0.85 and mean spatial
error of 0.125 on normal data. Under fault injection, the three
representative subsets --- low-light (Fig.~\ref{fig:fault_lowlight}),
rain (Fig.~\ref{fig:fault_rain}), and fog/footprint
(Fig.~\ref{fig:fault_fog}) --- show $R^2_\text{Overall}$ ranging
from 0.755 to 0.840, RMSE from 0.180 to 0.209, and MAE from 0.120
to 0.156 across all fault scenarios. While the $R^2$ values under
fault injection remain surprisingly high relative to the normal
baseline --- reflecting the model's retained coarse spatial awareness
--- the within-0.10 localization accuracy drops sharply from the
normal-data mean spatial error region to a range of only 0.310--0.445
across fault conditions, confirming that fine-grained steering
precision is substantially more sensitive to visual degradation than
coarse directional prediction.
The worst-performing fault scenarios include \texttt{FOG\_SLIGHT\_015}
and \texttt{FOG\_VARIABLE\_010} ($R^2\!\approx\!0.755$,
RMSE\,$\approx\!0.209$) and \texttt{RAIN\_PARTIAL\_003}
($R^2\!\approx\!0.768$, RMSE\,$\approx\!0.206$), while the least
disruptive conditions such as \texttt{RAIN\_OCCLUSION\_SMALL\_007}
and \texttt{FOG\_SIMULATION\_007} retain RMSE values as low as
0.180--0.181, highlighting significant variance in fault severity
across degradation families. The within-0.20 accuracy remains
relatively stable across fault types (0.658--0.752), indicating
that the model preserves coarse-grained directional steering under
most synthesized degradations.
These results collectively confirm that normal-data performance alone
is insufficient to guarantee robustness in real-world edge AI
deployment, motivating the need for fault-aware training strategies
and domain-specific augmentation pipelines.

\begin{table*}[htbp]
\caption{ResNet-18 Lane-Following Performance: Normal vs.\ Fault-Injected Data}
\begin{center}
\begin{tabular}{|l|c|c|c|c|c|}
\hline
\textbf{Condition} & \textbf{R\textsuperscript{2}} & \textbf{MSE}
    & \textbf{RMSE} & \textbf{Within-0.10} & \textbf{Within-0.20} \\
\hline
Normal (baseline)                               & $\sim$0.85 & $\sim$0.011 & $\sim$0.105 & ---    & ---           \\
\hline
Fault range (3 subsets)                         & 0.755--0.840 & ---       & 0.180--0.209 & 0.31--0.45 & 0.66--0.75 \\
\hline
Best fault (\texttt{\small FOG\_SIMULATION\_007})  & $\sim$0.835 & ---      & $\sim$0.181 & $\sim$0.445 & $\sim$0.752 \\
\hline
Worst fault (\texttt{\small FOG\_SLIGHT\_015})     & $\sim$0.755 & ---      & $\sim$0.209 & $\sim$0.310 & $\sim$0.662 \\
\hline
\textbf{Max degradation vs.\ baseline}         & \textbf{$-$11\%} & ---  & \textbf{$+$99\%} & --- & --- \\
\hline
\multicolumn{6}{l}{$^{\mathrm{a}}$Metrics computed on lane-following test data across three fault categories.}
\end{tabular}
\label{tab:comparison}
\end{center}
\end{table*}

\section{Conclusion}
\label{sec:conclusion}
We presented a decoupled framework for AI-driven safety testing of 
edge-deployed autonomous systems, where LLMs and LDMs execute offline 
to synthesize semantically rich fault scenarios, with results stored in 
lightweight lookup tables for efficient onboard queries. Using the 
VisionFault dataset of different fault categories spanning atmospheric, 
optical, sensor, motion, and illumination degradations, evaluation of 
our ResNet-18 lane-following model revealed substantial performance drops 
under fault conditions, with $R^2_\text{Overall}$ declining from 
${\sim}0.85$ to as low as $0.755$, RMSE increasing by up to 99\% (from 
${\sim}0.105$ to ${\sim}0.209$), and within-0.10 localization accuracy 
falling to as low as $31.0\%$ under the most severe fog scenarios. The 
within-0.20 coarse steering accuracy remained relatively stable 
($0.658$--$0.752$), indicating the model retains directional awareness 
while losing fine-grained precision. These findings expose critical 
vulnerabilities in edge AI perception and underscore that normal-data 
performance alone is insufficient for safe deployment. In future work, 
we will compare LLM--LDM-generated fault scenarios with randomly generated 
ones to quantify the benefits of semantically guided fault injection for 
robust lane-following evaluation.

\section*{Acknowledgment}
This paper is the result of preliminary work by Hamm-
Lippstadt University of Applied Sciences, Germany on the
project This work was supported by the EdgeAI-Trust project
"Decentralized Edge Intelligence: Advancing Trust, Safety,
and Sustainability in Europe".

\bibliographystyle{IEEEtran}
\bibliography{ref}
\end{document}